\documentclass[10pt,journal]{IEEEtran}

\usepackage{amsmath}
\usepackage{amssymb}
\usepackage{bm}
\usepackage{graphicx}
\usepackage{psfrag}
\usepackage{subfigure}
\usepackage{caption}
\usepackage{color}
\usepackage{comment}
\usepackage[colorlinks=true]{hyperref}
\usepackage[dvipsnames]{xcolor}
\usepackage{calrsfs}
\usepackage{paralist}

\usepackage{enumitem}
\usepackage{relsize}
\usepackage{tabularx}

\DeclareMathAlphabet{\pazocal}{OMS}{zplm}{m}{n}

\newcommand{\dps}{\displaystyle}

\newcommand{\Fone}{\pazocal{F}^{[1]}}
\newcommand{\Ftwo}{\pazocal{F}^{[2]}}

\newcommand{\Fn}{\pazocal{F}^{[n]}}
\newcommand{\fone}{f^{[1]}}
\newcommand{\ftwo}{f^{[2]}}

\newcommand{\fn}{f^{[n]}}
\newcommand{\mone}{\mu^{[1]}}
\newcommand{\mtwo}{\mu^{[2]}}

\newcommand{\mn}{\mu^{[n]}}

\DeclareMathOperator*{\argmax}{arg\,max}

\newtheorem{remark}{Remark}

\makeatletter
\newcommand{\subalign}[1]{%
  \vcenter{%
    \Let@ \restore@math@cr \default@tag
    \baselineskip\fontdimen10 \scriptfont\tw@
    \advance\baselineskip\fontdimen12 \scriptfont\tw@
    \lineskip\thr@@\fontdimen8 \scriptfont\thr@@
    \lineskiplimit\lineskip
    \ialign{\hfil$\m@th\scriptstyle##$&$\m@th\scriptstyle{}##$\crcr
      #1\crcr
    }%
  }
}
\makeatother

\ifCLASSOPTIONcompsoc
\usepackage[nocompress]{cite}
\else
 \usepackage{cite}
\fi

\ifCLASSINFOpdf
\else
\fi

\makeatletter
\newcommand{\bigger}{\bBigg@{4}}
\makeatother

\allowdisplaybreaks
\begin{document}

\setlength{\abovedisplayskip}{3pt}
\setlength{\belowdisplayskip}{3pt}

\setlength{\textfloatsep}{4pt plus 1.0pt minus 2.0pt}

% Reducing the spacings in equations (get a more compact equation representation)
\setlength{\medmuskip}{0mu}
\setlength{\thickmuskip}{0mu}
\setlength{\thinmuskip}{0mu}

%%%%%%%%%%%%%%%%%%%

\newcommand{\papertitle}{\huge{A note on the potentials of probabilistic and fuzzy logic}\\ 
%\vspace*{1ex}
%\Large {Potentials of probability and fuzziness for handling real-life uncertainties}
\vspace*{2ex}}
%%%%%%%%%%%%%%%%%%%%%%%%%%%%%%%%%%%%%

\title{\papertitle}

\author{Anahita~Jamshidnejad
		
                \thanks{\IEEEcompsocthanksitem A.\ Jamshidnejad
                   is with the Department of Control and Operations, 
                   Faculty of Aerospace Engineering,  
                   Delft University of Technology, 
                   }  \thanks{* Corresponding author:
                   \href{mailto:a.jamshidnejad@tudelft.nl}{a.jamshidnejad@tudelft.nl}.}
%\thanks{Manuscript received ???, 2018.}
}

\markboth{ }
{Shell \MakeLowercase{\textit{et al.}}: Bare Demo of IEEEtran.cls for IEEE Journals}

%%%%%%%%%%%%%%%%%%%%%%%%%%%%%%                  ^------------^------------^----Do not want these spaces!

\maketitle

\begin{abstract}

%% introducing control architecture

%%%%%%%%%
% Contributions
This paper mainly focuses on (1) a generalized treatment of fuzzy sets of type $n$, 
where $n$ is an integer larger than or equal to $1$, 
with an example, mathematical discussions, and real-life interpretation of the 
given mathematical concepts;  
(2) the potentials and links between fuzzy logic and probability logic that have not been 
discussed in one document in literature; 
(3) representation of real-life random and fuzzy uncertainties and ambiguities that arise in data-driven 
real-life problems, due to uncertain mathematical and vague verbal terms in datasets.    
\end{abstract}

\begin{IEEEkeywords}
Type-$n$ fuzzy sets; fuzzy logic; probability theory.%
\end{IEEEkeywords}

\IEEEpeerreviewmaketitle

\section{Introduction}
\label{sec:introduction}

%%%%%%%%%%%%%%%%%%%%
%\IEEEPARstart{}

%\label{sec:contributions}
%
%\noindent
In this paper an extensive treatment of type-2 fuzzy sets and membership functions is presented 
that is more general than the ones that can be found in literature.   
Moreover, the potentials of fuzzy logic and probability theory are discussed, 
as well as how real-life random and fuzzy uncertainties may be represented mathematically. 
%%%%%%%%%
\noindent
The paper is organized as follows:
Section~\ref{sec:intro:FL} provides an extensive discussion about fuzzy logic and the potential 
of this logic in representing linguistic uncertainties.
Section~\ref{sec:probabilistic_logic} discusses the probability theory.  
Section~\ref{sec:probability_versus_fuzziness} represents various formulations of 
random and fuzzy uncertainties.%
\begin{comment}  
Table~\ref{table:mathematical_notations}
gives the frequently-used mathematical notations. Note that throughout this paper 
capital letters for sets and small  
letters for functions are used. A subscript $n$ is used to indicate the type of 
a fuzzy set or membership function.%

%\bgroup
%\def\columnstretch{1}

\setlength{\tabcolsep}{5pt}
\begin{table}
\caption{Frequently-used mathematical notations.}
\label{table:mathematical_notations}
\begin{tabularx}{\linewidth}{l|X}
\hline
$x$ & variable\\
\hline
\end{tabularx}
\end{table}
\end{comment}

%%%%%%%%%%%%%%%%%%%%%%%%%%%%%%%%%%%%%%%%%%%%%%%%%%%%

%%%%%%%%%%%%%%%%%%%%%%%%%%%%%%%%%%%%%%%%%%%%%%%%%%%

\section{Fuzzy logic}
\label{sec:intro:FL}

Two main concepts that mathematical logic deals with are \emph{sets} and \emph{propositions} \cite{Shoenfield:1967}. 
Fuzzy logic was introduced by Zadeh in the 1960s \cite{Zadeh:1965,Zadeh:1968} to 
extend the concepts of set theory and propositional calculus, which by then were 
analyzed only through classical logic. 
Fuzzy logic is a continuous multi-valued logic system. fuzzy logic may be 
considered as a generalization to the classical logic.
 Linguistic variable, a concept unique to fuzzy logic, allows 
 this logic to serve as a basis for computations based on verbal information
\cite{Zadeh:2008,Zadeh:1975,Zadeh:1975-2}. 
Moreover, inspired by the procedure of reasoning of humans upon imprecise information, FL maps an imprecise concept into one with a higher precision \cite{Zadeh:1999}.%

A set in classical logic is a crisp concept: a mathematical object 
(e.g., a number, partition, matrix, variable, ...) 
%the realized value of a variable $x$ 
either ``belongs to'' the set, i.e., the degree of membership 
of the mathematical object to the set is $1$, or ``does not belong to" the set, 
i.e., the degree of membership of the mathematical object to the set is $0$. 
Therefore, a crisp set $\pazocal{C}$ may be expressed as a collection of 
mathematical objects, e.g., 
\begin{align}
\label{eq:crisp_set} 
\pazocal{C} = \left\{x_1,x_2, \ldots , x_n \right\}.
%,\quad x_i\in\mathbb{R}^{r\times c},\ i = 1,2, \ldots, n. 
\end{align}
%with $x$ a real-valued $p \times q$ mathematical variable varying within $\pazocal{R}_x 
%\subseteq \mathbb{R}^{p \times q}$. 
Similarly, a proposition in classical logic is either ``true'' 
(may be quantified by a crisp value $1$) 
or ``false'' (may be quantified by a crisp value $0$).% 
 
In fuzzy logic, sets are fuzzy concepts \cite{Zimmermann:1996}. 
Our main focus is on the general case of type-$n$ fuzzy sets, with 
$n=1, 2, 3, \ldots$. 
To motivate this, we start with an example.%

\subsection*{Opening example}

\begin{figure}
\centering
\psfrag{age}[][][.8]{Age}
\psfrag{MD}[][][.8]{Membership degree}
\psfrag{PMD}[][][.8]{Primary membership degree}
\psfrag{PMD3}[][][.8][14]{\hspace*{5ex} Primary membership degree}
\psfrag{SMD}[][][.8]{Secondary membership degree}
\psfrag{a3}[][][.8][-25 ]{Age}
\psfrag{40}[][][.7]{$40$}
\psfrag{0}[][][.7]{$0$}
\psfrag{1}[][][.7]{$1$}
\psfrag{9}[][][.7]{\hspace*{-2.5ex}$0.9$}
\psfrag{983}[][][.7]{$0.98$}
\psfrag{98}[][][.7]{\hspace*{-50ex}$0.98$}
\psfrag{57}[][][.7]{\hspace*{-3.8ex}$0.57$}
\psfrag{82}[][][.7]{$0.82$}
\psfrag{83}[][][.7]{\hspace*{-5ex} {\color {red} $0.83$}}
\psfrag{88}[][][.7]{{\color{red}$0.88$}}
\psfrag{27}[][][.7]{$27$}
\psfrag{y}[][][.8]{young}
\includegraphics[width = .55\linewidth]{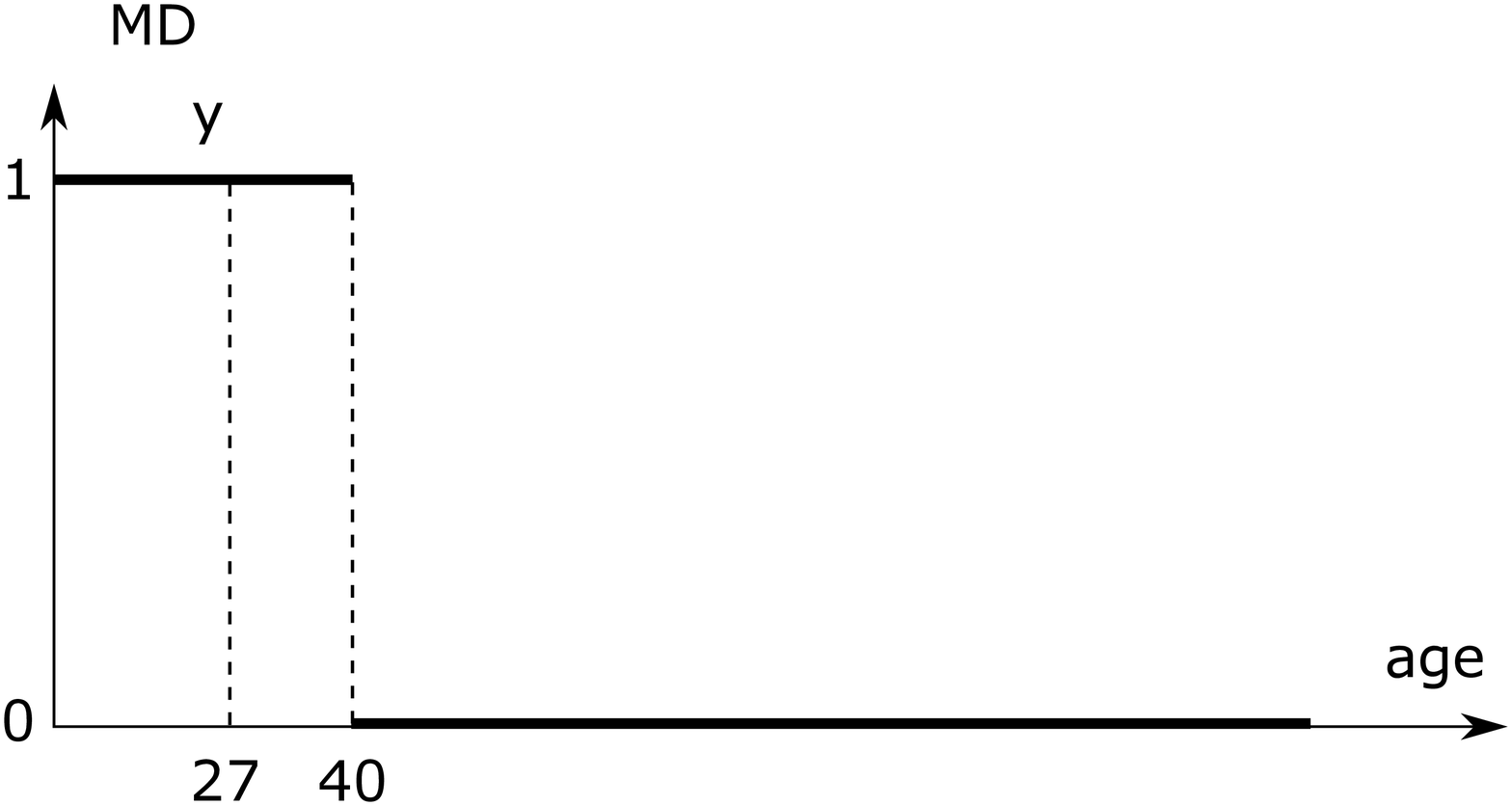}
\caption{Using crisp sets for quantifying \emph{young}.}
\label{fig:young_crisp_set}
\vspace*{4ex}

\includegraphics[width = .55\linewidth]{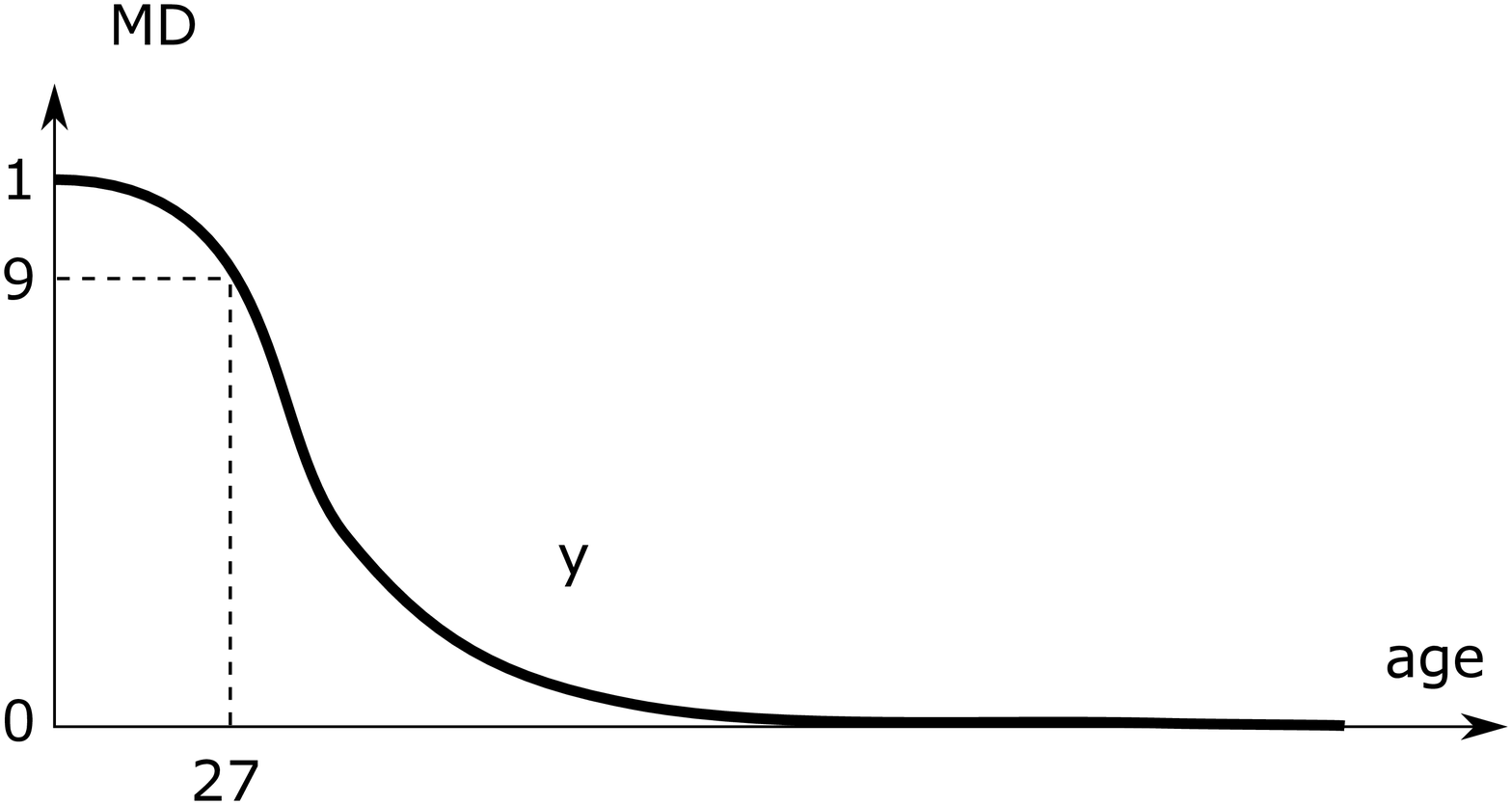}
\caption{Using type-$1$ fuzzy sets for quantifying \emph{young}.}
\label{fig:young_type1_fuzzy_set}
\vspace*{4ex}

\includegraphics[width = .55\linewidth]{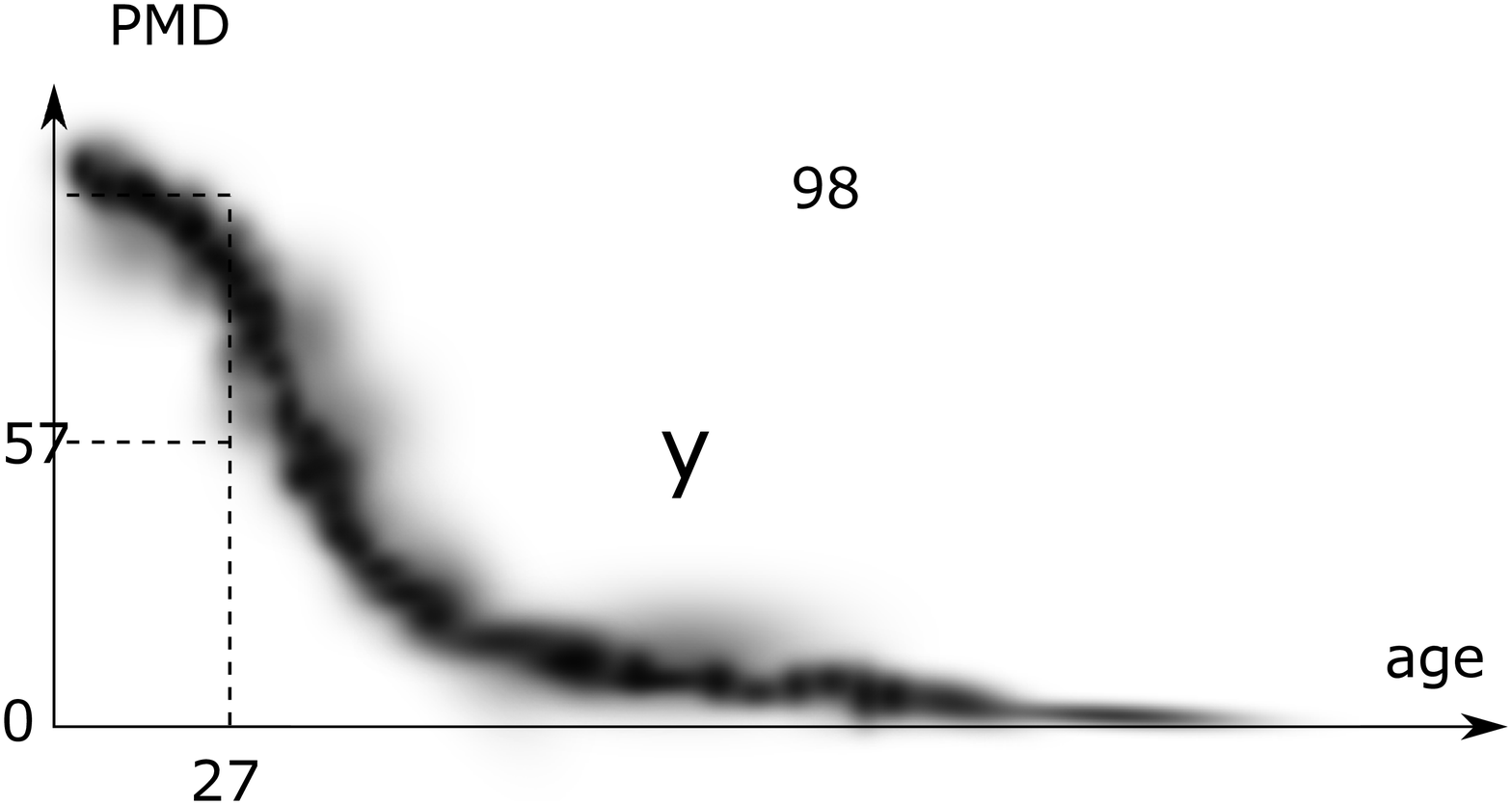}
\caption{Using type-2 fuzzy sets for quantifying \emph{young}.}
\label{fig:young_type2_fuzzy_set}
\vspace*{2ex}

\includegraphics[width = .65 \linewidth]{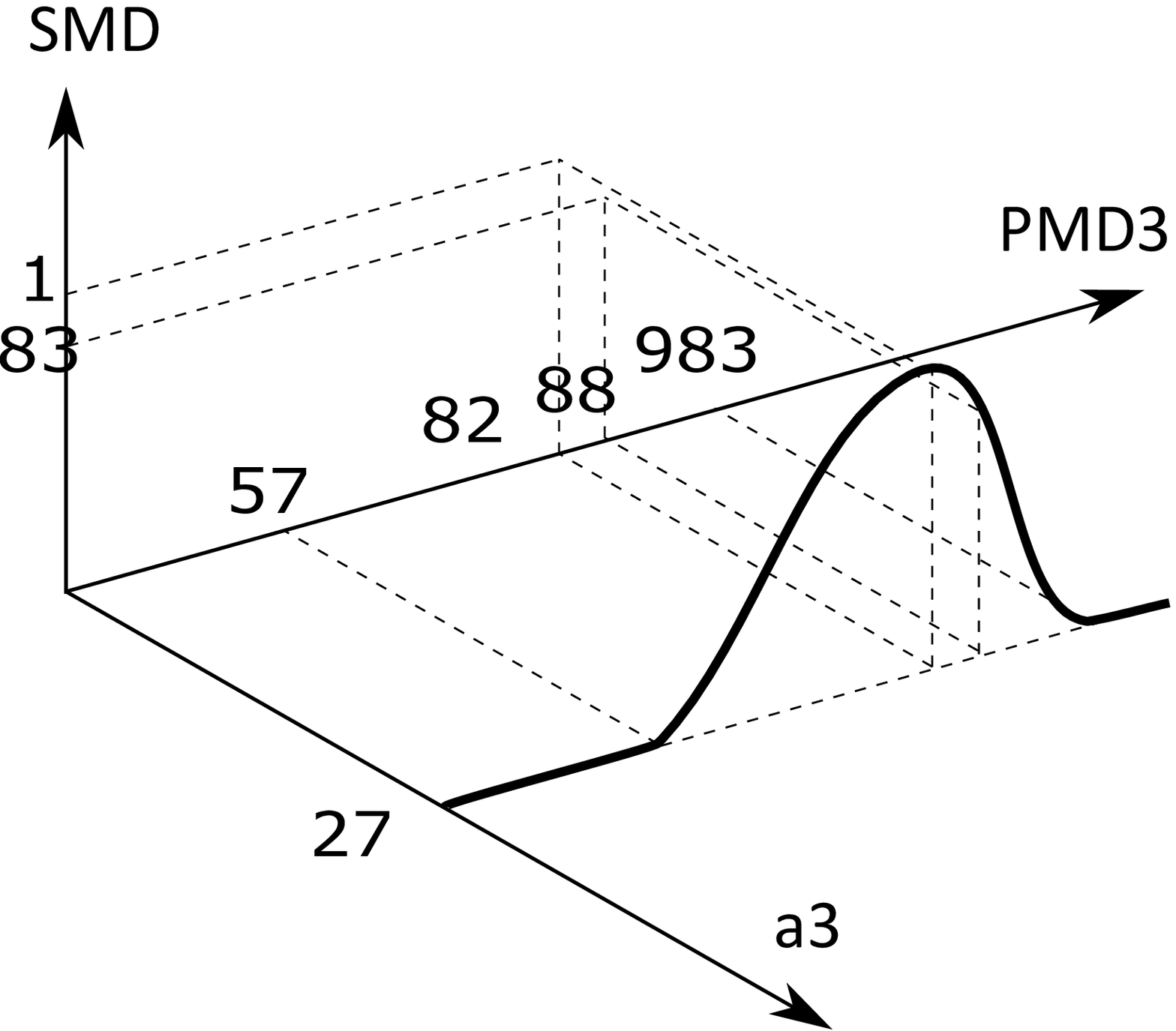}
\caption{3D representation of the type-2 membership function for 
quantifying \emph{young}, represented for the specific age of 27.}
\label{fig:young_3D_type_2_MF}
\vspace*{2ex}
\end{figure}
$\rhd$
Suppose that based on the information ``Felix is 27'', we would like to 
answer to the question: \emph{(To what extent) 
is Felix young?} 
Since \emph{young} is a qualitative concept, while $27$ is quantitative, 
these concepts should be bridged by quantifying \emph{young} and corresponding 
the quantitative information about Felix's age with the quantified 
definition of \emph{young}.%

One may define \emph{young} with the graph represented in Figure~\ref{fig:young_crisp_set}, 
which corresponds to a definition that categorizes people in two crisp sets, 
\emph{young} and \emph{not young}, i.e., people at an age below $40$  
belong to the set \emph{young} and otherwise they are \emph{not young}. 
Then, in quantitative terms, ``Felix belongs to the set 
\emph{young} with a membership degree of $1$'', and in qualitative terms, 
``Felix is \textbf{certainly} young''.% 

Alternatively, \emph{young} may be quantified via the graph shown in Figure~\ref{fig:young_type1_fuzzy_set}, 
where instead of \emph{certainly young} and \emph{certainly old}, ages vary  
in a spectrum, i.e., the degree of membership to the set 
\emph{young} varies in $[0,1]$ instead of $\{0 , 1\}$. 
Then, quantitatively, ``Felix  belongs to the set \emph{young} with a 
membership degree of $0.9$''. Qualitatively, ``Felix is \textbf{to a high 
extent} young''. \emph{Young} is then quantified using a type-$1$ fuzzy set.% 

Next, suppose that the border of the curve that defines \emph{young} is not strictly known.  
Figure~\ref{fig:young_type2_fuzzy_set} is an example where the intensity of the 
color black indicates our certainty about where any point at the border of 
the separating curve may lie in the given 2-dimensional plane.  
By just looking at the vertical axis (which is called the \emph{primary} membership degree), 
one may reply ``Felix belongs to the set \emph{young} with a \emph{primary} membership 
degree in the interval $[0.57,0.98]$'', or 
``Felix \textbf{may to a high extent} be young''.   
Now consider a third dimension that quantifies the intensity 
of the color black with real values between $0$ and $1$, i.e., black corresponds 
to $1$ and white corresponds to $0$ (see figure~\ref{fig:young_3D_type_2_MF}). 
This dimension represents the \emph{secondary} membership degree. Then one can respond 
``Felix is young with a \emph{primary} membership degree varying in the interval 
$[0.57, 0.98]$, and a \emph{secondary} membership degree varying in the interval 
 $[0,1]$. 
 For instance, ``Felix's age corresponds to a secondary membership degree of $0.83$ 
 for the primary membership degree of $0.88$''. 
 A type-$2$ fuzzy set (see \cite{Liang:1999,Mendel:2014,Hagras:2007_2} for more details on this type of fuzzy sets) has been used to quantify \emph{young} for this case.% 

More generally, if one interprets the word \emph{young} with a curve with 
 blurry borders that extend in $3$ (instead of $2$) dimensions, 
 i.e., the graph of young can be re-illustrated in $4$ dimensions considering 
 \emph{primary}, \emph{secondary}, and \emph{tertiary} membership degrees, the 
 qualitative word young has actually been quantified using a type-$3$ fuzzy set. 
 This concept can further be extended to type-$n$ fuzzy sets for 
 which the corresponding membership functions can be illustrated in 
 $n + 1$ dimensions, and there are $n$ membership degrees involved. $\lhd$%
\vspace*{-2ex}

\subsection*{Mathematical Discussion}
  
Next, the concept of type-$n$ fuzzy sets will be detailed mathematically. 
In contrast to a crisp set, a mathematical object belongs 
to any type-$n$ fuzzy set, with $n=1,\,2,\,\ldots$, with 
 a primary, secondary, \ldots membership degree varying in 
 $[0,1]$.  
Similarly, in fuzzy logic a proposition may ``to a certain extent'' be true or false. 
To link the two theories of fuzzy propositions and fuzzy sets, one may consider 
two fuzzy sets, the ``set of true propositions'' and the ``set of false propositions''. 
Any new proposition, which corresponds to one or several mathematical variables  
called membership degrees, can belong to both these sets at the same time with  
certain degrees of membership.%
 
Generally, a membership function $ f^{[n]}: \textrm{dom}\left( \fn \right) \rightarrow [0,1]$ 
together with its domain, $\textrm{dom}\left( \fn \right)$, which itself can be a fuzzy 
set, characterizes a fuzzy set. 
For a type-1 fuzzy set, 
\begin{align}
\label{eq:fuzzy_set_type1}
\Fone=\bigg\{
\Big(x_1,\mone_1\Big),
\Big(x_2,\mone_2\Big),
\ldots,
\Big(x_n,\mone_n\Big)
\bigg\},
\end{align}
the corresponding type-1 membership function\footnote{A 
membership function is named after the corresponding fuzzy set, 
i.e., they both are of the same type.} 
$\fone$ generates the degree of membership of mathematical objects $x_i$   
within $\pazocal{C}$ to the fuzzy set $\Fone$, 
i.e., 
\begin{align}
\label{eq:fuzzy_membership_function_type1}
\fone: \Big\{x_1, x_2, \ldots , x_n \Big\} 
\rightarrow [0,1]: \ 
x_i \mapsto \mone_i
\end{align}
or equivalently,
\begin{align}
\label{eq:fuzzy_membership_function_type1_eq}
\mone_i = \fone(x_i),\quad i = 1, 2, \ldots, n.
\end{align}
In comparison with a crisp set, a type-1 fuzzy set $\Fone$  
is composed of pairs of mathematical objects and their degrees of 
membership to $ \Fone$ (see \eqref{eq:fuzzy_set_type1}).% 

\begin{remark}
One may generalize the above discussion by re-defining the 
crisp set $\pazocal{C}$ in \eqref{eq:crisp_set} as a special case of a type-1 
fuzzy set, where $\mone_i=1$, $i=1, 2, \ldots, n$. 
This means that  the corresponding type-1 
membership function of $\pazocal{C}$ is the unit function. 
This definition may even further be generalized (see the 
following discussions for more details), i.e., a crisp set may be 
re-defined as a type-$n$ fuzzy set, with $n = 1,2, \ldots$, with all 
the membership functions of order $1, \ldots, n$ the unit function.%  
\end{remark}
 
For data-driven approaches in engineering problems, large amount of information 
that are expressed in the human language, may be used to generate fuzzy sets.  
The degree to which a fuzzy set can handle the uncertainties and vagueness that 
exist in this information, depends on the type of the fuzzy set.  
The main difference between type-1 fuzzy sets and fuzzy sets of type 2 and larger 
is in the domain of their corresponding membership functions, i.e.,  
the domain of a type-1 membership function is a crisp set $\pazocal{C}$ 
(see \eqref{eq:fuzzy_set_type1}), or equivalently, a type-1 fuzzy set 
with the unit function as its membership function, while the domain of a type-$n$ 
fuzzy set with $n = 2, 3, \ldots$ is a union of various fuzzy sets of type $n-1$.%

More specifically, a type-2 fuzzy set $\Ftwo$,
\begin{align}
\label{eq:fuzzy_set_type2}
\Ftwo = 
\Bigg\{
\bigg(\Big(x_1,\mone_{1,1}\Big),\mtwo_{1,1} \bigg),
&\bigg(\Big(x_1,\mone_{2,1}\Big),\mtwo_{2,1} \bigg),
\nonumber\\
\ldots, 
&
\bigg(\Big(x_1,\mone_{m_1,1}\Big),\mtwo_{m_1,1} \bigg),
\nonumber\\
&\vdots
\nonumber\\
\bigg(\Big(x_n,\mone_{1,n}\Big),\mtwo_{1,n} \bigg),
&\bigg(\Big(x_n,\mone_{2,n}\Big),\mtwo_{2,n} \bigg),
\nonumber\\
\ldots, 
&
\bigg(\Big(x_n,\mone_{m_n,n}\Big),\mtwo_{m_n,n} \bigg)
\Bigg\}, 
\end{align} 
corresponds to a type-2 membership function $\ftwo$, which generates the 
secondary degree of membership of any mathematical object $\Big(x_i,\mone_{j,i}\Big)$ 
within the type-1 fuzzy set $\pazocal{C} \times \pazocal{R}$ to the type-2 fuzzy set $\Ftwo$, 
where $\pazocal{R} \subseteq [0,1]$, 
and $i=1, \ldots, n$, $j = 1, \ldots, m_i$. 
Therefore, 
\begin{align}
\label{eq:fuzzy_membership_function_type2}
\ftwo : 
\bigg\{
&\Big(x_1,\mone_{1,1}\Big),
\Big(x_1,\mone_{2,1}\Big),
\ldots, 
\Big(x_1,\mone_{m_1,1}\Big),
\nonumber\\
&\ldots
\nonumber\\
&\Big(x_n,\mone_{1,n}\Big),
\Big(x_n,\mone_{2,n}\Big),
\ldots, 
\Big(x_n,\mone_{m_n,n}\Big)
\bigg\}\rightarrow
[0,1]:
\nonumber\\
&\left( x_i, \mone_{j,i} \right) \mapsto \mtwo_{j,i}, 
\end{align} 
or equivalently,
\begin{align}
\label{eq:fuzzy_membership_function_type2_eq}
\mtwo_{j,i} = \ftwo\Big(\left( x_i, \mone_{j,i} \right)\Big),
\qquad i=1,\ldots,n,\quad j=1,\ldots, m_i.
\end{align}
We can reformulate the domain of $\ftwo$ as:
\begin{align}
\textrm{dom}\left( \ftwo \right) := 
\bigg\{
&\Big(x_1,\mone_{1,1}\Big),\
\ldots, 
\Big(x_n,\mone_{1,n}\Big)
\Big\}
\cup 
\\
\bigg\{
&\Big(x_1,\mone_{2,1}\Big),\
\ldots, 
\Big(x_n,\mone_{2,n}\Big)
\Big\}
\cup 
\nonumber\\
&\vdots 
\nonumber\\
\bigg\{
&\Big(x_1,\mone_{m_1,1}\Big),\
\ldots, 
\Big(x_n,\mone_{m_1,n}\Big)
\Big\}
\cup 
\nonumber\\
&\vdots
\nonumber\\
\bigg\{
&\Big(x_k,\mone_{m_k,1}\Big)
\Big\} 
= \bigcup_{i=1}^n
  \bigcup_{j=1}^{m_i}
\left\{
\left( x_i, \mone_{j,i} \right)
\right\}
,
\nonumber
\end{align}
assuming that 
\[
\dps\argmax_i\ \{ m_i \} = k. 
\]
This shows the domain of the type-$2$ membership function $\ftwo$ 
is the union of $m_k$ type-$1$ fuzzy sets. 
Therefore, a type-$2$ fuzzy set also corresponds to $m_k$ 
type-$1$ membership functions $\fone_j$ that generate the primary membership degrees 
$\mone_{j,i}$, i.e., 
\begin{align}
\label{eq:primary_membership_functions}
\mone_{j,i} = \fone_j(x_i),\qquad i = 1, \ldots , n, \quad j = 1, \ldots , m_i.
\end{align}
The domains of these type-$1$ membership functions are $\pazocal{C}$ or 
a subset of $\pazocal{C}$.% 
  
In summary, to identify a type-$2$ fuzzy set $\Ftwo$ on a domain $\pazocal{C}$ 
of the independent variables $x_i$, one needs to know the corresponding type-$2$ 
membership function $\ftwo$ and the domain of this function. 
Equivalently, one should know $\ftwo$ and the $m_k$ type-$1$ membership functions 
$\fone_j$. 
From \eqref{eq:fuzzy_set_type2}, a type-2 fuzzy set $\Ftwo$ is composed of pairs, 
where each pair itself includes a pair consisting of the independent variable $x_i$ 
and a primary membership degree $\mone_{j,i}$ of $x_i$, and a secondary membership 
degree.%

%Similarly, a type-3 fuzzy set $\pazocal{F}_3$, ranged over the union of $m_2$ 
%type-2 fuzzy sets, is defined by:
%\begin{align}
%\label{eq:fuzzy_set_type3}
%\pazocal{F}_3 = \bigg\{
%\Big\{
%\pazocal{R}_x \times \pazocal{D}_{\mu_1 , i}, \quad 
%i = 1, 2, \ldots &, m_1
%\Big\}
%\times \pazocal{D}_{\mu_2 , j}, 
%\\
%&j = 1 , 2, \ldots , m_2
%\bigg\}
%\times 
%\pazocal{D}_{\mu_3}.  \nonumber
%\end{align}  

Generally speaking, a type-$n$ fuzzy set $\Fn$ includes 
mathematical objects of the form:
\[
\Bigg(
\ldots
\bigg( 
\left( x_i , \mone_{j_{n-1},\ldots,j_1,i} \right), \mtwo_{j_{n-1},\ldots,j_1,i}
\bigg)
\ldots
\bigg)
\mn_{j_{n-1},\ldots,j_1,i}
\Bigg)
\] 
with $j_1 \in \left\{1, \ldots, m_{i,1}\right\}$, \ldots,  
$j_{n - 1} \in \left\{1, \ldots, m_{i,n -1}\right\} $, and corresponds to a type-$n$ membership 
function $\fn$, which generates the $n^{\textrm{th}}$ membership degrees $
\mn_{j_{n-1},\ldots,j_1,i} $ of any mathematical object that belongs to the union 
of $m_{k_n}$ fuzzy sets of type $n - 1$, with 
\[
k_n = \dps \argmax_i\{ m_{i , n - 1} \}.
\] 
A type-$n$ fuzzy set corresponds to $\argmax_i\{ m_{i , n - 1} \}$ 
membership functions of type $n - 1$, 
to $\argmax_i\{ m_{i , n - 2} \} $ membership functions of type $n - 2$, 
\ldots, and to $\argmax_i\{ m_{i , 1} \}$ type-$1$ membership functions.  
One should know all these membership functions, as well as the unique type-$n$ 
membership function to identify $\Fn$.%

%*****************************************************************
\vspace*{-2ex}

\subsection*{Interpretation for real-life problems}
\label{sec:physical_interpretation}

Based on the discussions given above, the following conclusions can be made. 
For an independent variable $x_i \in \pazocal{C}$, the \emph{extent} that it belongs to 
a type-$1$ fuzzy set $\Fone$ is characterized by one uncertainty, which is specified by the 
degree of membership of $x_i$ to $\Fone$.%

Any type-$2$ fuzzy set $\Ftwo$ corresponds to $m_k$ type-$1$ membership 
functions and hence, the $m_k$ corresponding type-$1$ fuzzy sets. 
Therefore, $\forall x_i \in \pazocal{C}$, the \emph{extent} that $x_i $ belongs to 
$\Ftwo$ is characterized by two uncertainties:
\begin{compactitem}
\item
Uncertainty about the extent to which $x_i$ belongs to the union or either of the 
$m_k$ corresponding type-$1$ fuzzy sets. This is quantified by the primary  
membership degrees. 
\item
Uncertainty about the extent to which $x_i$, which belongs with specific 
primary membership degrees to the type-$1$ fuzzy sets, belongs to $\Ftwo$. 
This is quantified  by the secondary membership degree. 
\end{compactitem}%

Generally speaking, the \emph{extent} that $x_i \in \pazocal{C}$ belongs to a 
type-$n$ fuzzy set $\Fn$ is characterized by $n$ uncertainties:
\begin{compactitem}
\item
Uncertainty about the extent to which  $x_i$  belongs  to the union or either of  the 
corresponding $\argmax_i\{ m_{i , 1} \}$ type-$1$ fuzzy sets. 
This is quantified by the primary membership degrees.
\item[\vdots]
\item
Uncertainty about the extent to which $x_i$ belongs to the union or either of the 
corresponding $\argmax_i\{ m_{i ,n-1} \}$ fuzzy sets of type $n-1$. 
This is quantified by the $(n-1)^\textrm{th}$ membership degrees.
\item
Uncertainty about the extent to which $x_i$ belongs to $\Fn$.      
This is quantified by the $n^\textrm{th}$ membership degree.
\end{compactitem}%%

%$\forall x^* \in \pazocal{R}_2$ (with 
%$\pazocal{R}_2$ the range of the type-2 membership function), 
%$x^*$ is ``to some extent'' in the set, where ``to some extent'' should be quantified by 
%$\mu_2 \in \pazocal{D}_{\mu_2}$. Moreover, $\forall x \in \pazocal{R}_x$, $x$ is ``to some 
%extent'' in $\pazocal{R}_2$, where ``to some extent'' may be quantified by  $\mu_{1,1} \in \pazocal{D}_{1,1}$ 
%or\footnote{This is a ``logical or'', mathematically indicated by $\lor$.} \ldots 
%or $\mu_{1,m_1} \in \pazocal{D}_{1,m_1}$. Hence, there are two uncertainties in this case:
%\begin{compactitem}
%\item
%\emph{To what extent (mathematically quantified) does $x$ belong to $\pazocal{R}_2$?}  
%\item
%\emph{To what extent (mathematically quantified) does $x$ belong to $\pazocal{F}_2$?} 
%\end{compactitem}%
%
%In summary, for a generally type-$n$ fuzzy set ($n \geq 2$) $\pazocal{F}_n$, 
%there are $n$ uncertainties involved:
%\begin{compactitem}
%\item
%\emph{To what extent (mathematically quantified) does $x$ belong to $\pazocal{R}_n$?}  
%\item
%\emph{To what extent (mathematically quantified) does $x$ belong to $\pazocal{R}_{n - 1}$?}
%\item[] \vdots
%\item
%\emph{To what extent (mathematically quantified) does $x$ belong to $\pazocal{R}_2$?}
%\item
%\emph{To what extent (mathematically quantified) does $x$ belong to $\pazocal{F}_n$?}  
%\end{compactitem}%

%%%%%%%%%%%%%%%%%%%%%%%%%%%%%%%%%%%%%%%%%%%%%%%%%%%%%%%%%%%%%%%%%%%%%%%%%%%%%%%%%%%%%%%%%%
\section{Probabilistic logic}
\label{sec:probabilistic_logic}

\begin{figure}
\psfrag{T}[][][1]{True set $\pazocal{T}$}
\psfrag{F}[][][1]{False set $\pazocal{F}$}
\psfrag{W}[][][1]{World set  $\pazocal{W}$}
\psfrag{u}[][][.8]{\hspace*{38ex} {\color{black}$\pazocal{T} \cap \pazocal{F}$}}
\psfrag{n}[][][.8]{\hspace*{54ex} {\color{black}$\pazocal{T} \backslash \pazocal{F}$}}
\psfrag{p}[][][.8]{ \hspace*{25ex} {\color{black}$\pazocal{F} \backslash \pazocal{T}$} }
\includegraphics[width = \linewidth]{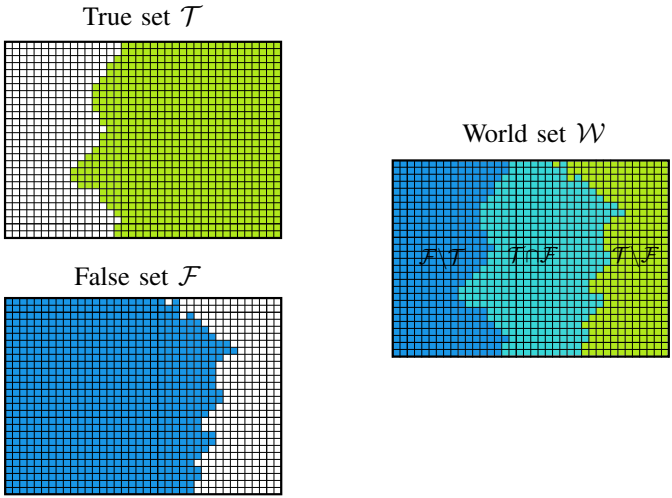}
\caption{The sets of true $\pazocal{T}$ and false $\pazocal{F}$ propositions in fuzzy 
logic may have an overlap, 
i.e., $\pazocal{T} \cap \pazocal{F} \neq \emptyset$. 
Moreover,some mathematical objects that belong to $\pazocal{F} \backslash \pazocal{T}$ 
or $\pazocal{T} \backslash \pazocal{F}$ may not necessarily have a membership degree 
of $1$.} 
\label{fig:fuzzy_T_F_overlap}
\end{figure}

In this section, probabilistic logic \cite{Nilsson:1986} versus fuzzy logic is discussed. 
With the analogy explained before, a proposition in mathematical logic     
may correspond to a mathematical object that can belong to either  
(crisp or fuzzy) sets of true or false propositions.   
For the sake of brevity, the discussions of this 
section are represented using the concept of sets only. To notify the correspondence of sets 
and propositions, the notations $\pazocal{T}$ (referring to \emph{true}) 
and $\pazocal{F}$ (referring to \emph{false})  
are used for the sets that build up the possible world\footnote{Possible world of 
the event $E$ is a world set that embraces all possible realizations of $E$.}, 
$\pazocal{W}$, of an event $E$.%

\begin{figure}
\psfrag{T}[][][1]{True set $\pazocal{T}$}
\psfrag{F}[][][1]{False set $\pazocal{F}$}
\psfrag{W}[][][1]{World set  $\pazocal{W}$}
\psfrag{f}[][][.8]{\hspace*{50ex} {\color{white}$\pazocal{T}$}}
\psfrag{t}[][][.8]{ \hspace*{25ex} {\color{white}$\pazocal{F}$} }
\includegraphics[width = \linewidth]{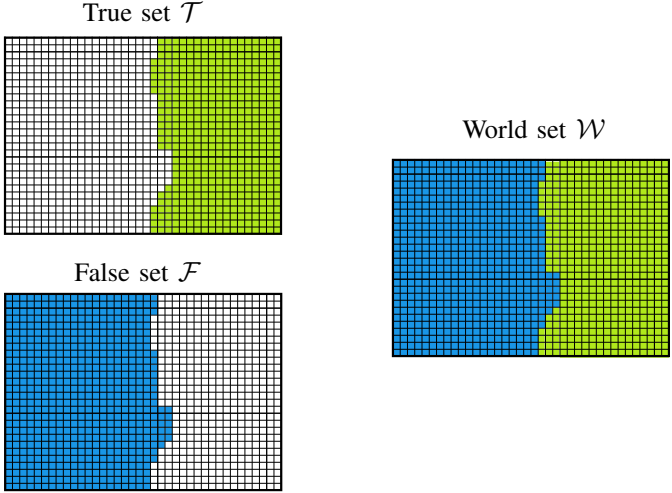}
\caption{The sets of true $\pazocal{T}$ and false $\pazocal{F}$ propositions in probabilistic logic 
do not have any overlap, and any mathematical object that does not belong 
to either $\pazocal{F}$ or $\pazocal{T}$, belongs to the other set with 
a probability of $1$.}
\label{fig:probabilistic_T_F_no_overlap}
\end{figure}

A main difference between fuzzy logic and probabilistic logic is that for the former, 
$\pazocal{F}$ and $\pazocal{T}$ are fuzzy sets, while for the latter, they are crisp sets. 
Therefore, the borders of $\pazocal{F}$ and $\pazocal{T}$ in fuzzy logic are exposed to 
uncertainties. Hence, there may be an overlap between these two sets, i.e., 
we may have $\pazocal{T} \cap \pazocal{F} \neq \emptyset$ (see Figure~\ref{fig:fuzzy_T_F_overlap}), 
while in probabilistic logic we necessarily have 
$\pazocal{T} \cap \pazocal{F}= \emptyset$ (see figure~\ref{fig:probabilistic_T_F_no_overlap}).%
 
Any realization of the repeatable event $E$ can correspond to a mathematical object $x$. 
In probabilistic logic, in case in $100$ experiments, where $E$ is repeated, for 
$\Pi_1$ experiments $x$ goes to $\pazocal{F}$ and for $\Pi_2$ experiments $x$ goes to 
$\pazocal{T}$, then $\Pi_1 + \Pi_2 =100$. The normalized values $\pi_1$ and $\pi_2$ of 
the natural values $\Pi_1$ and $\Pi_2$ are called the probability of $x$ belonging to 
 $\pazocal{F}$ and $\pazocal{T}$, respectively. From $100$ experiments in fuzzy logic, 
 if in $M_1$ experiments $x$ goes to $\pazocal{F}$ and in $M_2$ experiments $x$ goes to 
 $\pazocal{T}$, then $M_1 + M_2$ is not necessarily $100$ (may be larger than 
 $100$).       
The reason is that for some experiments $x$ may go to the overlap $\pazocal{T} \cap \pazocal{F}$, 
i.e., the experiment is doubly counted in both $M_1$ and $M_2$. 
%For some experiments, $x$ may go to either $\pazocal{F} \backslash \pazocal{T}$  
%or $\pazocal{T} \backslash \pazocal{F}$, but with a membership degree smaller than $1$ 
%(see the less intense colored parts of $\pazocal{F} \backslash \pazocal{T}$ and 
%$\pazocal{T} \backslash \pazocal{F}$ in Figure~\ref{fig:fuzzy_T_F_overlap}). 
%Therefore, a real value smaller than $1$ may be added to either $M_1$ or $M_2$. 
The normalized values $\mu_1$ and $\mu_2$ of the real values $M_1$ and $M_2$ 
are called the degrees of membership of $x$ to the fuzzy sets $\pazocal{F}$ 
and $\pazocal{T}$, respectively.%

To generalize, the summation $\sum_{i=1}^{s_\textrm{c}} \pi_i$ of the probabilities $\pi_i$ 
of $x$ belonging to $s_\textrm{c}$ crisp sets that build up the possible world $\pazocal{W}$ 
of the event $E$ is necessarily $1$, while the sum $\sum_{i=1}^{s_\textrm{f}} \mu_i$ of the 
degrees of membership of $x$ to $s_\textrm{f}$ fuzzy sets that build up $\pazocal{W}$ 
may be smaller than, equal to, or larger than $1$.%    

%%%%%%%%%%%%%%%%%%%%%%%%%%%%%%%%%%%%%%%%%%%%%%%%%%%%%%%%%%%%%%%%%%%%%%%%%%%%%%%%%%%%%%%%%%

\section{ Mathematical formulation of uncertainties in real-life events}
\label{sec:probability_versus_fuzziness}

In real life, an event $E$ may be prone to various types of imprecision 
and hence, uncertainties.% 

\subsection*{Uncertainties before realization of an event}

Suppose that the world set $\pazocal{W}$  of possible realizations of an event $E$ is known, 
i.e., we know and can measure (or estimate the value of) all the characteristics 
that can define and distinguish any specific position within $\pazocal{W}$ 
(this is called a \emph{state} in real-life engineering problems).%

\begin{figure}
\psfrag{W}[][][1]{$\pazocal{W}$}
\includegraphics[ width = \linewidth ]{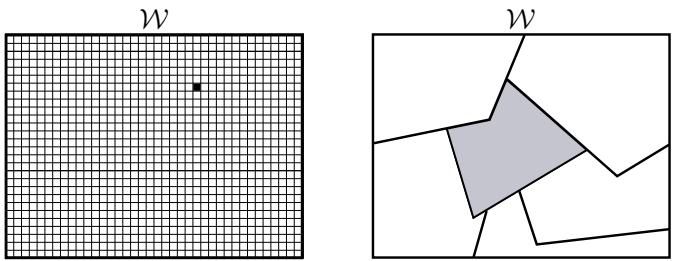}
\caption{When the realization of an event is measurable, it will take an exact position 
(illustrated by the black dot in the left-hand side plot) in the world set. 
When the realization of an event is non-measurable, it will take position in 
a subset (colored in grey in the right-hand side plot) of the world set, while 
its exact position is uncertain. The corresponding subsets are defined to 
make the non-measurable realization, precise in meaning. 
Although the subsets illustrated in this plot are crisp, the may in general be fuzzy.}
\label{fig:measurable_and_nonmeasurable_realizations}
\end{figure}

Before $E$ is realized, there is always uncertainty about which possible realization 
is going to occur. If the realization of $E$ is measurable (e.g., 
a temperature, which can be measured directly using a thermostat), the realization 
is \emph{precise in value}, is determined by the same measurable characteristics as those 
that define the world set, and hence its position in the world set can strictly be distinguished 
(see the left-hand side plot in Figure~\ref{fig:measurable_and_nonmeasurable_realizations}).  
Therefore, the uncertainty is one-fold, i.e., uncertainty about 
the possible measurements from the world set that will be realized in occurrence of $E$.
In this case, if one knows and can measure (or estimate the value of) all the effective 
factors (these are called the controllable and uncontrollable inputs in real-life 
engineering problems) that may play a role in any realization of $E$ 
(or equivalently in any realization of the characteristics that define 
the world set $\pazocal{W}$), then probabilistic logic can determine the probability 
of occurrence  of any possible realization. 
In short, when all the concepts and characteristics involved in the procedure of 
realization of an event are measurable or \emph{precise in value}, then probabilistic 
logic can handle the uncertainties.% 
 
In case the realization of $E$ is non-measurable (e.g., the comfort, which cannot 
be measured directly using a measurement device), this realization is \emph{imprecise 
in value}. Then the uncertainty is two-fold, i.e., uncertainty about which 
possible non-measurable realization in the world set is going to occur, and uncertainty about 
the exact position of the realization in the world set.
Systems in real life work with measurable concepts. For instance, the non-measurable 
instruction ``when the room's comfort is low, decrease the temperature'', 
should be transformed into a measurable instruction for an air conditioning system, 
e.g., ``when the room's temperature is between $23$ and $25$ degrees of Celsius, 
decrease the temperature for $3$ degrees of Celsius''. 
This transforms a concept that is \emph{imprecise in value} into one that 
is \emph{precise in meaning}, and reduces the original uncertainty about the 
exact position of the concept in the world set $\pazocal{W}$ to an uncertainty about 
the exact position of the concept in a known subset of $\pazocal{W}$.
Fuzzy logic can transform a realization that is imprecise in value into 
one that is precise in meaning by assigning a membership function (of, 
generally, type $n$) to the characteristics that are imprecise in value.%  

%\begin{figure}
%\psfrag{s}[][][.85]{state}
%\includegraphics[width = \linewidth]{MD_T_F_W_unionTF}
%\end{figure}

It is important to note that although the transformation from a realization 
that is imprecise in value into one that is precise in meaning is with the 
aim of reducing the uncertainty from the entire world set to a subset of it,  
further uncertainties may or may not exist in the exact position of the borders 
of these subsets. 
For instance, when a specific type of search-and-rescue robot is deployed to 
a burning building, the domains of temperature for which the robot is functional 
and dysfunctional should be determined. 
Knowing the materials, sensors, ... used in the construction of the robot, 
one can determine the temperature at which this robot or any 
other robot of this type will become dysfunctional.    
In this case, the subsets functional and dysfunctional are crisp.
In this case the transformation to a realization that is precise in meaning 
using membership functions (fuzzy logic) is identical to the transformation 
using probability functions (probabilistic logic). 
Distinguishing the exact borders of these two subsets for a human search-and-rescuer 
is more challenging, since humans are not as \emph{homogeneous} as identical robots 
produced in a factory. Therefore, these borders may vary from person to person and 
the resulting subsets of functional and dysfunctional can become fuzzy. 
Then the only right tool for transforming the non-measurable realization 
into one that is precise in meaning is fuzzy logic.%     

\subsection*{Uncertainties after realization of an event}

After $E$ is realized, in case the realization of $E$ is measurable, e.g., 
a temperature, which can be measured directly using a thermostat, the realization 
is \emph{precise in value}  
and its position in the world set can strictly be distinguished. 
When the realization of $E$ is non-measurable, e.g., the comfort, which cannot 
be measured directly using a measurement device, this realization is imprecise 
in value. Therefore, there is uncertainty about its position in the world set. 
The realization should first be quantified to become precise in meaning, and 
only then, a subset of the world set that embeds all the possible positions 
of the realization in the world set, together with the degree to which the realization 
is positioned at any of these possible positions can be determined.% 

\bibliographystyle{ieeetr}
\bibliography{Reference_list_editted}

\end{document}